\documentclass[twoside,twocolumn,9pt]{article}
\usepackage{extsizes}
\usepackage[super,sort&compress,comma]{natbib} 
\usepackage[version=3]{mhchem}
\usepackage[left=1.5cm, right=1.5cm, top=1.785cm, bottom=2.0cm]{geometry}
\usepackage{balance}
\usepackage{mathptmx}
\usepackage{sectsty}
\usepackage{graphicx} 
\usepackage{lastpage}
\usepackage[format=plain,justification=justified,singlelinecheck=false,font={stretch=1.125,small,sf},labelfont=bf,labelsep=space]{caption}
\usepackage{float}
\usepackage{fancyhdr}
\usepackage{fnpos}
\usepackage[english]{babel}
\addto{\captionsenglish}{%
  
}
\usepackage{array}
\usepackage{droidsans}
\usepackage{charter}
\usepackage[T1]{fontenc}
\usepackage[usenames,dvipsnames]{xcolor}
\usepackage{setspace}
\usepackage[compact]{titlesec}
\usepackage{hyperref}
\usepackage{algorithm}
\usepackage{algpseudocode}
\usepackage{afterpage}
\usepackage{placeins}

\usepackage{epstopdf}

\definecolor{cream}{RGB}{222,217,201}

\begin{document}

\pagestyle{fancy}
\thispagestyle{plain}
\fancypagestyle{plain}{
\renewcommand{\headrulewidth}{0pt}
}

\makeFNbottom
\makeatletter
\renewcommand\LARGE{\@setfontsize\LARGE{15pt}{17}}
\renewcommand\Large{\@setfontsize\Large{12pt}{14}}
\renewcommand\large{\@setfontsize\large{10pt}{12}}
\renewcommand\footnotesize{\@setfontsize\footnotesize{7pt}{10}}
\makeatother

\renewcommand{\thefootnote}{\fnsymbol{footnote}}
\renewcommand\footnoterule{\vspace*{1pt}%
\color{cream}\hrule width 3.5in height 0.4pt \color{black}\vspace*{5pt}} 
\setcounter{secnumdepth}{5}

\makeatletter 
\renewcommand\@biblabel[1]{#1}            
\renewcommand\@makefntext[1]%
{\noindent\makebox[0pt][r]{\@thefnmark\,}#1}
\makeatother 
\renewcommand{\figurename}{\small{Fig.}~}
\sectionfont{\sffamily\Large}
\subsectionfont{\normalsize}
\subsubsectionfont{\bf}
\setstretch{1.125} 
\setlength{\skip\footins}{0.8cm}
\setlength{\footnotesep}{0.25cm}
\setlength{\jot}{10pt}
\titlespacing*{\section}{0pt}{4pt}{4pt}
\titlespacing*{\subsection}{0pt}{15pt}{1pt}

\fancyfoot{}
\fancyfoot[RO]{\footnotesize{\sffamily{1--\pageref{LastPage} ~\textbar  \hspace{2pt}\thepage}}}
\fancyfoot[LE]{\footnotesize{\sffamily{\thepage~\textbar\hspace{2pt} 1--\pageref{LastPage}}}}
\fancyhead{}
\renewcommand{\headrulewidth}{0pt} 
\renewcommand{\footrulewidth}{0pt}
\setlength{\arrayrulewidth}{1pt}
\setlength{\columnsep}{6.5mm}
\setlength\bibsep{1pt}

\makeatletter 
\newlength{\figrulesep} 
\setlength{\figrulesep}{0.5\textfloatsep} 

\newcommand{\topfigrule}{\vspace*{-1pt}%
\noindent{\color{cream}\rule[-\figrulesep]{\columnwidth}{1.5pt}} }

\newcommand{\botfigrule}{\vspace*{-2pt}%
\noindent{\color{cream}\rule[\figrulesep]{\columnwidth}{1.5pt}} }

\newcommand{\dblfigrule}{\vspace*{-1pt}%
\noindent{\color{cream}\rule[-\figrulesep]{\textwidth}{1.5pt}} }

\makeatother

\twocolumn[
  \begin{@twocolumnfalse}
\vspace{1em}
\sffamily
\begin{tabular}{m{1cm} p{16cm} m{1cm}}

& \noindent\LARGE{\textbf{Automated Retrosynthesis Planning of Macromolecules Using Large Language Models and Knowledge Graphs}} & \\
\vspace{0.3cm} & \vspace{0.3cm}  & \vspace{0.3cm}\\

 & \noindent\large{Qinyu Ma, \textit{$^{a}$} Yuhao Zhou,\textit{$^{a}$} Jianfeng Li$^\ast$\textit{$^{a}$} } \\

& \noindent\normalsize{

Identifying reliable synthesis pathways in materials chemistry is a complex task, particularly in polymer science, due to the intricate and often non-unique nomenclature of macromolecules. To address this challenge, we propose an agent system that integrates large language models (LLMs) and knowledge graphs (KGs). By leveraging LLMs' powerful capabilities for extracting and recognizing chemical substance names, and storing the extracted data in a structured knowledge graph, our system fully automates the retrieval of relevant literatures, extraction of reaction data, database querying, construction of retrosynthetic pathway trees, further expansion through the retrieval of additional literature and recommendation of optimal reaction pathways. A novel Multi-branched Reaction Pathway Search (MBRPS) algorithm enables the exploration of all pathways, with a particular focus on multi-branched ones, helping LLMs overcome weak reasoning in multi-branched paths.  This work represents the first attempt to develop a fully automated retrosynthesis planning agent tailored specially for macromolecules powered by LLMs. Applied to polyimide synthesis, our new approach constructs a retrosynthetic pathway tree with hundreds of pathways and recommends optimized routes, including both known and novel pathways, demonstrating its effectiveness and potential for broader applications.} \\

\end{tabular}

 \end{@twocolumnfalse} \vspace{0.6cm}

  ]

\renewcommand*\rmdefault{bch}\normalfont\upshape
\rmfamily
\section*{}
\vspace{-1cm}


\footnotetext{\textit{$^{a}$~The State Key Laboratory of Molecular Engineering of Polymers, Research Center of Al for Polymer Science, Department of Macromolecular Science, Fudan University, Shanghai 200433, China }}
\footnotetext{$\ast$~lijf@fudan.edu.cn}



\section{Introduction}
Retrosynthesis planning\cite{Segler2018} plays an important role in chemical engineering and chemistry research, offering a systematic approach to designing synthetic pathways for target compounds. By deconstructing complex molecules into simpler precursors, retrosynthesis enables researchers to navigate the vast possibilities of chemical transformations and efficiently plan synthesis routes. This process is also crucial for the advancement of material discovery, the optimization of chemical production, and the support of innovative research across disciplines. Current methods for single-step chemical retrosynthesis analysis primarily include computational approaches such as density functional theory for precise calculations\cite{Li2024} and using deep learning models for prediction. Deep learning-based prediction methods can be broadly categorized into template-based approaches\cite{Zhao2024,Chen2020} , which rely on predefined reaction templates for high precision but have limited applicability, and template-free approaches\cite{Lin2020,Karpov2019}, which offer greater flexibility but often sacrifice precision. However, these techniques primarily focus on decomposing target compounds into one intermediate and multiple starting molecules, leaving more complex multi-intermediate pathways largely unexplored. Moreover, research efforts have predominantly focused on small molecules, with limited attention to macromolecules such as polymers and proteins. 

The challenges in applying retrosynthesis planning to macromolecules are particularly noteworthy. Unlike small molecules, macromolecules often lack extensive, well-documented reaction databases, making the use of deep learning models for prediction tricky. Moreover, the large number of atoms in macromolecular systems, as well as the fact that chemical reactions are often influenced by complex interactions, make accurate calculations challenging. Therefore, researchers are often required to browse a large amount of academic papers for retrosynthesis planning of macromolecules. Unfortrunately, the extraction of reaction information from the literature and the construction of retrosynthetic pathways for macromolecules is further complicated by their complex and variable nomenclature, which makes traditional rule-based methods insufficient for accurately identifying relevant reactions. For instance, the polymer widely known as "polystyrene" may also appear as "Poly(1-phenylethylene)" based on structure-based naming or as "Poly(vinylbenzene)" and "Poly(ethenylbenzene)" under source-based conventions. To address these issues, more intelligent approaches are necessary.\cite{Hodge2020,Hu2021} A promising solution lies in leveraging LLMs to ensure the consistency of polymer material names, thereby enabling the construction of an entity-aligned knowledge graph\cite{Zeng2021} to facilitate the automated construction of retrosynthetic pathways. Despite the potential of this approach, no prior studies have investigated the integration of LLMs and knowledge graphs specifically for retrosynthesis planning. While Bran et al.\cite{Bran2024} previously utilized LLMs to automate aspects of chemistry research, their work treated retrosynthesis planning as a supporting tool, relying on underlying deep-learning methods for its implementation.

On the other hand, Large language models (LLMs)\cite{Mann2020,Radford2019} have reshaped natural language processing with their human-like text generation, complex pattern recognition, and adaptability to tasks including translation, summarization, and question answering\cite{Devlin2019}. Leveraging deep learning, they process vast amounts of text data\cite{Chalkidis2020,Yang2020}, proving invaluable for text mining\cite{Zhang2024,Chen2024}, research planning\cite{Bran2024,Boiko2023,liu2024toward}, and chemical applications\cite{Ramos2024}. Despite these advantages, LLMs face critical limitations.\cite{Bender2021,Srivastava2022} Their probabilistic nature\cite{Vaswani2017,Radford2018} and reliance on unverified data can lead to hallucinations\cite{Cao2021,Maynez2020}, while static datasets delay knowledge updates.\cite{Lewis2020} They also struggle with precise math, logic\cite{huang2022towards} and interpreting non-textual data like molecular structures or reaction schemes. In retrosynthesis planning\cite{Segler2018}, these limitations are particularly problematic, as the process requires accurate multi-step reaction predictions, real-time scientific knowledge, and the ability to assess pathway feasibility. Furthermore, LLMs lack the capability to generate structured outputs critical for mapping reaction networks. These challenges hinder their ability to reliably chart complex chemical pathways, especially for macromolecules.

To address these challenges, we propose a retrosynthesis planning agent based on large language models (LLMs) and knowledge graphs (KG) for materials chemistry. This agent is capable of automatically querying, downloading, and extracting chemical reaction information based on a given target product (see the demo video in SI). It then constructs a structured knowledge graph, facilitating efficient and accurate information retrieval and expansion. The agent utilizes a Memoized Depth-first Search (MDFS) algorithm\cite{Cormen2022,Tarjan1972,Knuth1997}, along with database queries, to construct a retrosynthetic pathway tree that synthesizes the target product using  commercially available compounds as starting compounds. When a reaction pathway cannot be further expanded, the agent automatically retrieves and incorporates additional synthesis data from relevant literature, continuously enriching the knowledge graph and further broadening and extending the chemical reaction pathways. Ultimately, with the help of the Multi-branched Reaction Pathway Search (MBRPS) Algorithm, the agent identifies all authoritative and feasible synthesis pathways, and recommends the optimal reaction pathway based on factors such as reaction conditions, yields and so on. The proposed approach is applied to polyimide synthesis, showcasing its ability to construct complex retrosynthetic pathway trees and recommend optimized routes, encompassing both established and novel pathways.

\section{Method}
\subsection{Automated Literature Retrieval}

\begin{figure}[h]
	\centering
	\includegraphics[width=0.95\linewidth]{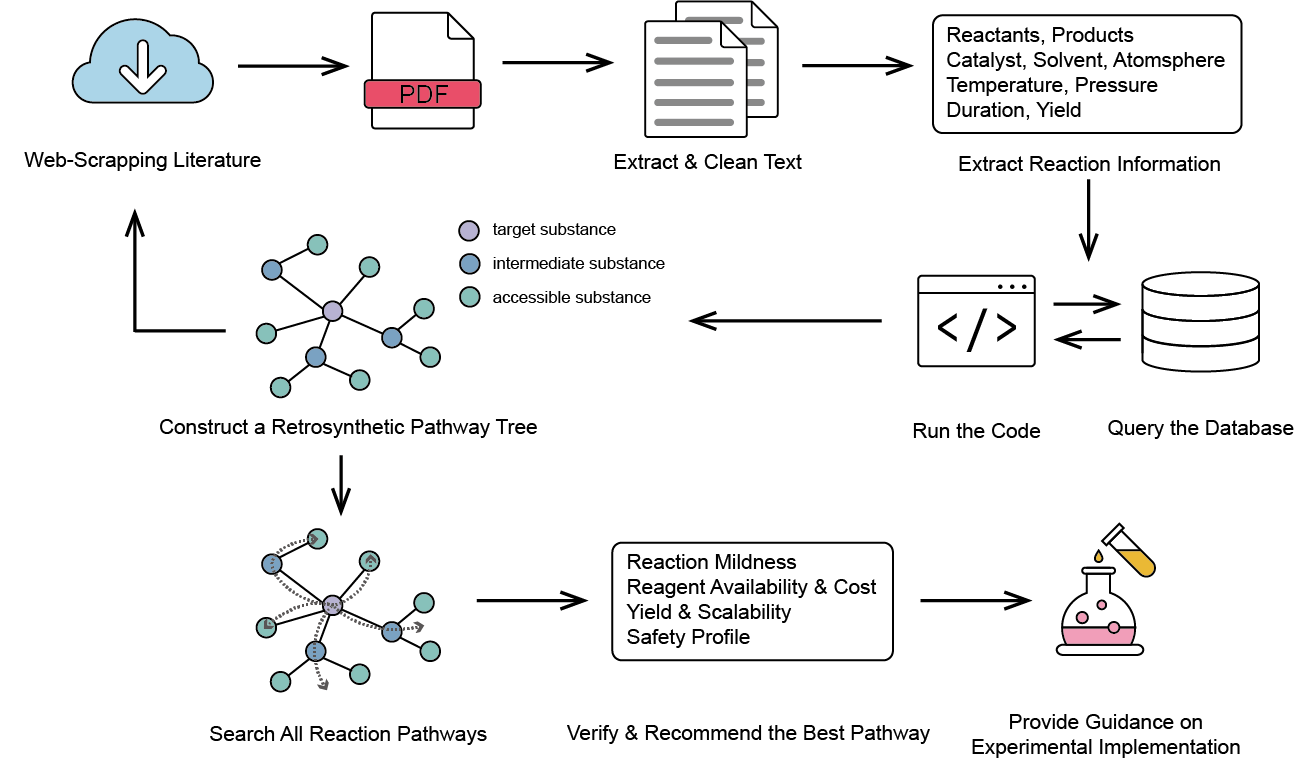}
	\caption{Schematic workflow for automated retrosynthesis planning using the LLM agent, covering literature retrieval, reaction data extraction, database querying, expansion and construction of retrosynthetic tree and optimal pathway recommendation.}
	\label{fig:fig1}
\end{figure}

The workflow of the automated retrosynthesis planning agent is illustrated in Fig.~\ref{fig:fig1}. A demo video in SI is also provided, showcasing the agent's ability to execute the entire workflow autonomously without any human intervention. The Agent first utilizes the Google Scholar API\cite{Cholewiak2021} to retrieve relevant paper titles based on predefined keywords. These titles are then used to download literature PDFs via web scraping. Text is extracted from the PDFs using PyMuPDF\cite{PyMuPDF2024}. Following extraction, the data is cleaned by removing special characters and symbols to enhance readability and ensure better comprehension by large language models (LLMs).

\begin{figure}[h]
	\centering
	\includegraphics[width=0.95\linewidth]{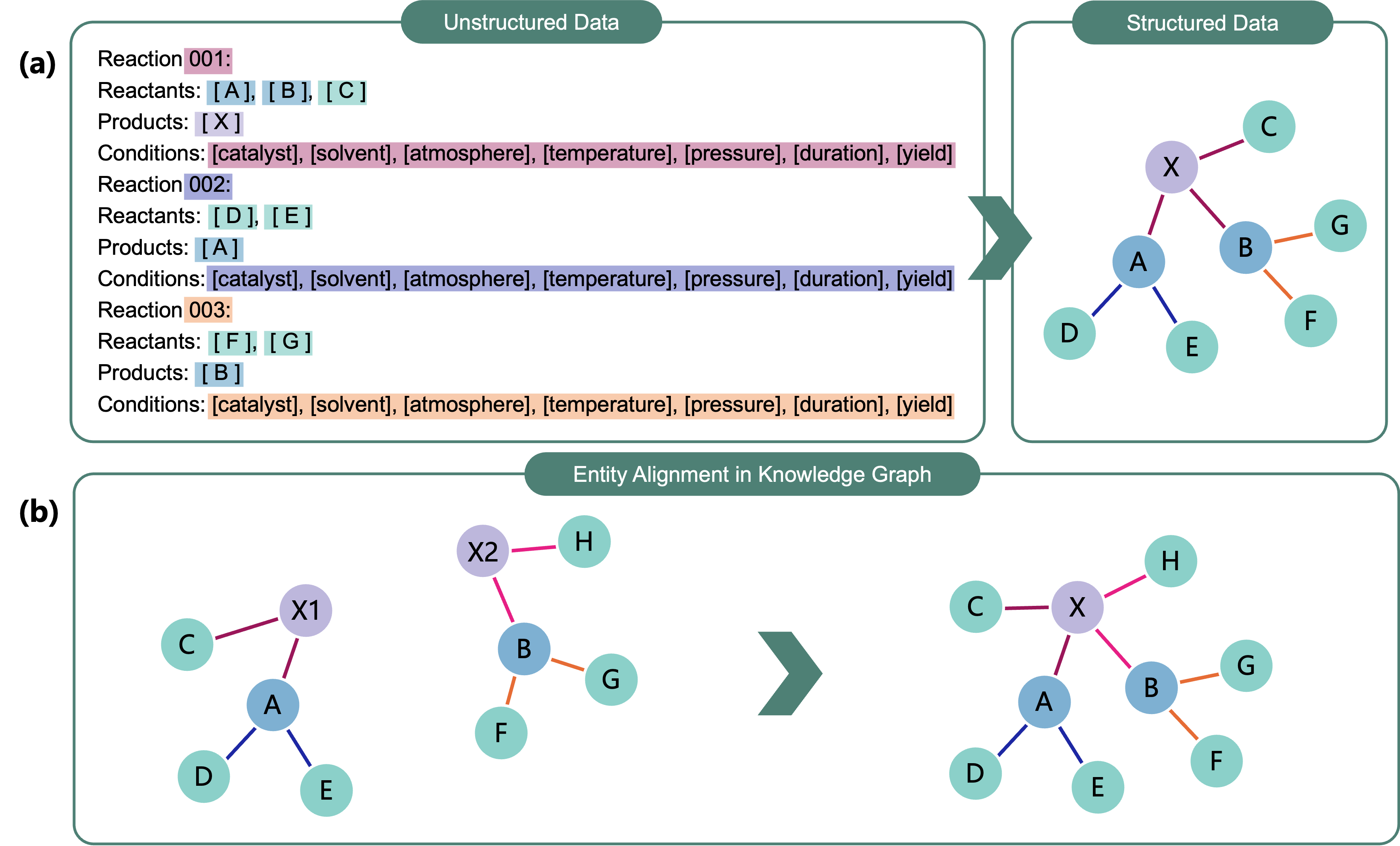}
	\caption{Schematic Diagram of Knowledge Graph Construction. (a) Schematic diagram for extracting chemical reactions and converting unstructured data into structured formats. (b) Entity alignment in knowledge graphs to ensure consistent naming across articles.}
	\label{fig:fig2}
\end{figure}

\subsection{Knowledge Graph Construction from Extracted Information}

Our agent is built upon the ChatGPT-4o API\cite{Achiam2023}. By utilizing this model, the agent employs prompt engineering\cite{Mann2020}, in-context learning\cite{Min2021} and Chain-of-Thought (CoT)\cite{Wei2022} to perform tasks including entity and relation extraction, knowledge graph construction, and entity alignment (Fig.~\ref{fig:fig2}). 

It processes the cleaned text and images to extract chemical reactions, which are then output in a standardized format. The extracted chemical reaction information includes the names of reactants and products, reaction temperature, pressure, catalysts, solvents, atmosphere, reaction duration, and yield. Leveraging ChatGPT-4o's inherent proficiency in text comprehension and standardized output, these tasks are effectively accomplished without the need for fine-tuning in most cases.\cite{Chen2024,Leong2024} However, there are occasional instances where non-reactive reagents are incorrectly listed as reactants, even with the use of prompting to constrain the model. We use the CoT technique to conduct a secondary verification to avoid this issue (See Section I of SI).

Based on the model’s structured outputs, the agent uses regular expressions to extract entities and relationships. Specifically, each reactant and product is treated as an entity, while reaction conditions, numerical identifiers and yields are considered as relations, forming unidirectional edges from reactants to products. Ultimately, the agent converts unstructured chemical reaction information from literature into a structured knowledge graph for efficient information retrieval and expansion.

Although  prompt engineering and in-context learning can ensure consistency in chemical substance names within a single paper, it is challenging to maintain consistency across multiple papers due to the input length limitations of LLMs. Therefore, the agent rechecks the knowledge graph to identify cases where different nodes represent the same substance. If such cases exist, the agent unifies them and updates the knowledge graph accordingly (See Section II of SI).

\subsection{Retrosynthetic Pathway Tree Construction and Expansion}

Utilizing the constructed knowledge graph, the agent employs a Memoized Depth-first Search (MDFS) algorithm\cite{Cormen2022,Tarjan1972,Knuth1997} to build the retrosynthetic pathway tree, with the target product as the root and leaf nodes representing commercially available compounds.

The goal of constructing a retrosynthetic pathway tree is to trace the reaction pathway step by step from the target substance back to the initial reactants. Specifically, each node in the retrosynthetic pathway tree represents a chemical substance, generated through a specific reaction. The construction of retrosynthetic tree follows a set of rules:

1. If the target substance is already present in the accessible set of initial reactants, it is marked as a leaf node, requiring no further expansion.

2. If the substance can be synthesized through any known reactions, it is considered expandable; otherwise, if it cannot be synthesized, it is considered non-expandable.

For expandable nodes, the MDFS algorithm traverses all reactions producing the substance, retrieves reactants one by one, and adds them as child nodes to the current node. To prevent cycles, the algorithm discards a path if the new node already exists in the set of parent nodes. This process is carried out recursively, ensuring the validity of each route. Ultimately, only valid reaction pathways are retained, while for nodes that cannot be further expanded or form a cycle, the corresponding reaction pathways will be completely removed, ensuring that the final tree structure accurately reflects the synthesis route from the target substance to the initial reactants.

During the recursive tree construction, the agent queries databases such as eMolecules\cite{eMolecules2024} and PubChem\cite{Kim2023}, along with additional commonly used polymers, to verify whether the current node represents a commonly uses substance. Additionally, the agent uses RDKit\cite{Bento2020} to convert the names of small molecules into standardized SMILES strings for database matching. If a node corresponds to a commonly used substance, it is designated as a leaf node, halting further expansion. To further enhance the efficiency of tree construction, a memory-augmented approach is employed, where the results of database queries, regarding whether a node substance corresponds to commonly used materials, are stored in a cache. This strategy eliminates the need for repeated database lookups of the same substance, significantly reducing computational overhead.

It is worth noting that not all reactants in a single paper are typically commercially available. Therefore, it is necessary to further investigate the literature on the synthesis of intermediate reactants, until commercially available compounds can be used to synthesize the intermediate. Similarly, during retrosynthesis tree construction, if a node cannot be further expanded to a leaf node, the agent will query the relevant literature on the synthesis of the intermediate corresponding to that node, extract relevant chemical reactions from it, and add them to the knowledge graph, thus helping the node successfully expand to a leaf node, enabling the construction of a complete reaction pathway. Ultimately, an expanded retrosynthetic pathway tree with the target substance as the root node is constructed, which includes multiple chemical reaction pathways that can synthesize the target substance from commercially available compounds.

\begin{figure}[h]
	\centering
	\includegraphics[width=0.95\linewidth]{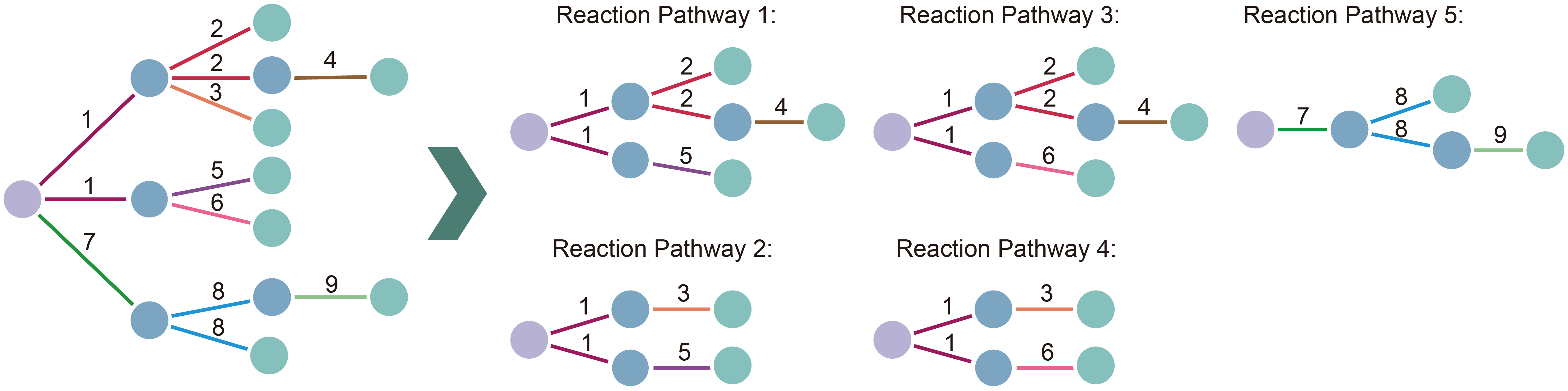}
	\caption{Reaction pathway searching in retrosynthetic pathway tree using Multi-branched Reaction Pathway Search (MBRPS) algorithm. In this typical example, all five reaction pathways have been identified.}
	\label{fig:fig3}
\end{figure}

\subsection{Chemical Reaction Pathways Search and Recommendation}

Upon constructing the retrosynthetic pathway tree for the target product, the agent employs the Multi-branched Reaction Pathway Search Algorithm (MBRPS) (Algorithm \ref{alg:MBRPS}), to identify all valid chemical reaction pathways, as illustrated in Fig.~\ref{fig:fig3}. This algorithm is specially designed for multi-branched reaction pathways, which are common in practical retrosynthesis planning (see Discussion section for more details). Each reaction pathway identified provides all the reactions required to synthesize the target product from commercially available compounds, with validation from a database. Specifically, a product can be synthesized through various reactions with corresponding reactants, each representing a node in the retrosynthetic pathway tree. For nodes associated with the same reaction index, they form an “AND” relationship, meaning that all child nodes with the same index must be included to synthesize the target compound. On the other hand, nodes with different reaction indices represent an “OR” relationship, indicating that the target compound can be synthesized by selecting one of several possible reactions. Based on this relationship, we use a recursive method to obtain all reaction paths for each node. This method identifies all valid synthesis routes, including multi-branched ones, helping LLMs overcome weak reasoning in multi-branched paths and enabling comprehensive exploration of reaction pathways.

\begin{algorithm}[h]
\caption{Multi-branched Reaction Pathway Search Algorithm}
\label{alg:MBRPS}
\small
\begin{algorithmic}[1]
\Function{SearchReactionPathways}{current node}
\Require Current node
\Ensure Reaction pathways as sequences of reaction indices
\If{current node is a leaf node}
    \State \Return an empty array
\EndIf
\State Initialize \texttt{PathwaysDict} to store reaction pathways for child nodes
\For{each child node of the current node}
    \State Obtain pathways by \Call{SearchReactionPathways}{child node}
    \State Get \texttt{ReactionIdx} for the child node
    \If{\texttt{ReactionIdx} is not in \texttt{PathwaysDict}}
        \State Add \texttt{ChildPaths} to \texttt{PathwaysDict} under \texttt{ReactionIdx}
    \Else
        \State Merge \texttt{ChildPaths} with existing pathways
    \EndIf
\EndFor
\State \Return pathways in \texttt{PathwaysDict} as an array
\EndFunction
\end{algorithmic}
\end{algorithm}

\FloatBarrier 
\begin{figure*}[ht]
	\centering
        \includegraphics[width=0.80\linewidth]{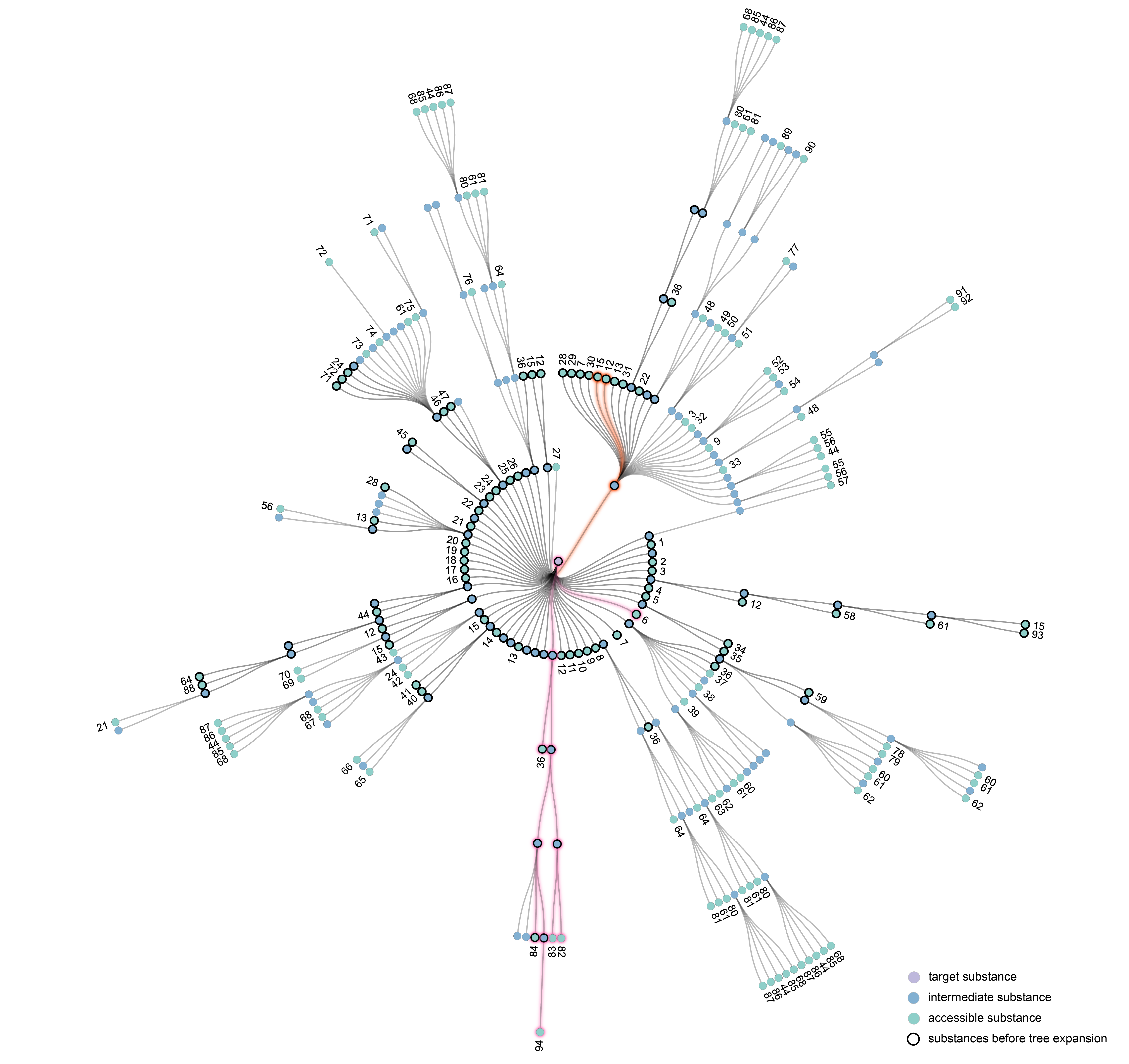}
        \includegraphics[width=0.85\linewidth]{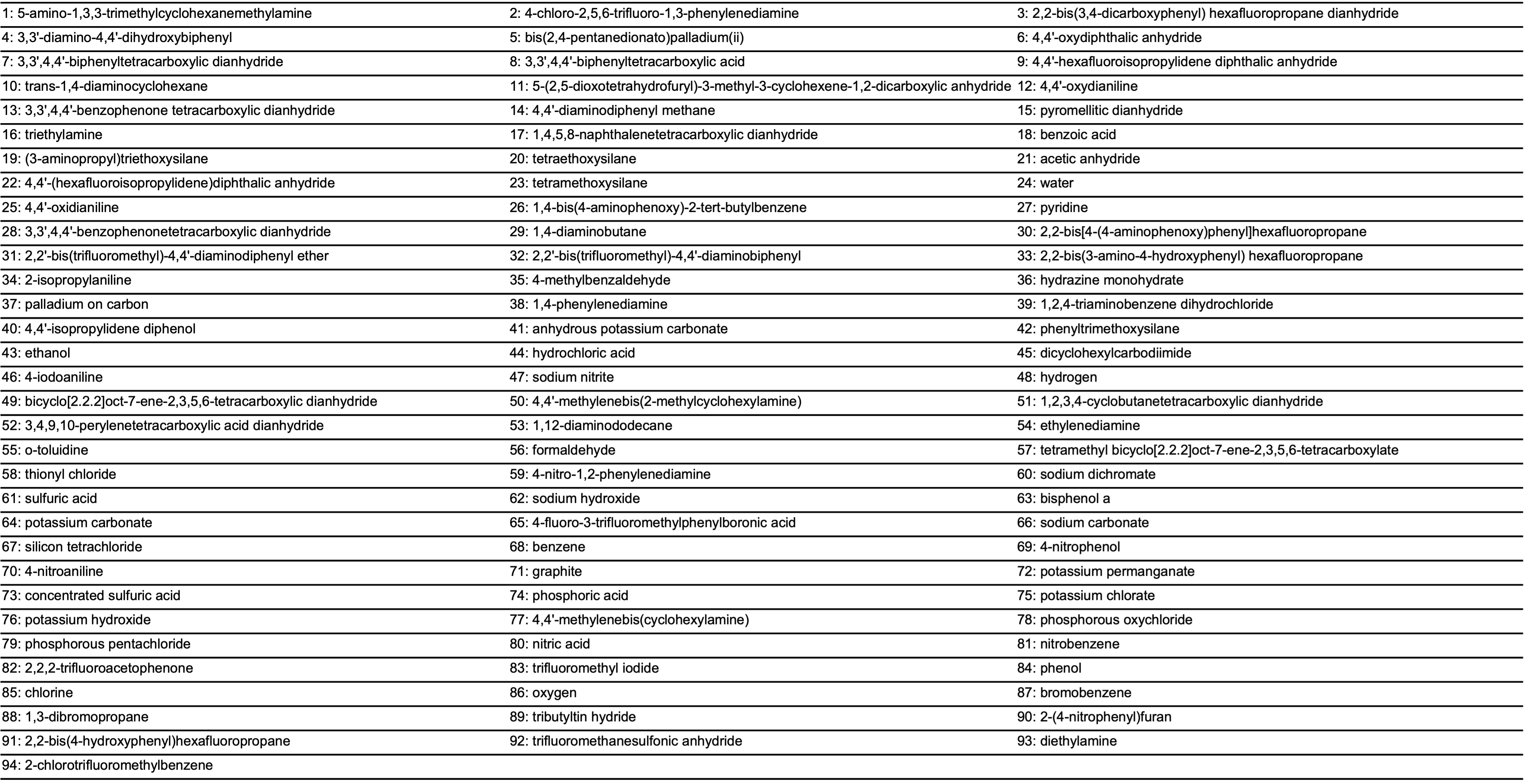}
	\caption{Expanded retrosynthetic pathway tree for polyimide based on 197 articles. For simplicity, duplicate child nodes with the same name at each node were hidden. The number of nodes before expansion was 322 (113 in the figure), and the number of nodes after expansion was 3099 (294 in the figure). The reaction path obtained based on criterion 1 (Fig.~\ref{fig:fig5}) is highlighted in orange. The reaction path obtained based on criterion 2 (Fig.~\ref{fig:fig6}) is highlighted in pink.}
	\label{fig:fig4}
\end{figure*}
\FloatBarrier 

Finally, the agent evaluates all identified pathways, considering various factors such as the availability and cost of reactants, catalysts, and solvents, the mildness of reaction conditions (e.g., low temperature, pressure, short duration), reaction yield and scalability, and the safety profile of reagents and conditions (e.g., toxicity, hazards), by leveraging Chain of Thought (CoT)\cite{Wei2022}. Based on these criteria, the agent recommends the optimal synthetic route for the target product, offering a more efficient and reliable solution for retrosynthesis planning.

\section{Results}
\subsection{Retrosynthetic Pathway Tree of Polyimide}

The aforementioned method is applied to Polyimide (PI), a high-performance polymer renowned for its exceptional thermal stability, chemical resistance, and mechanical strength. These properties make PI indispensable in industries such as aerospace, electronics, and high-temperature applications.\cite{Li2024b} However, its complex synthesis and high production costs have driven research into optimizing its synthetic routes.\cite{Huang2023} Therefore, we have chosen polyimide for retrosynthesis pathway analysis to explore more efficient and cost-effective approaches. By designating "polyimide" as the target substance for retrosynthetic analysis, the agent retrieved 39 research papers on polyimide synthesis methods, extracting chemical reactions from these sources, and converted them into a structured knowledge graph format, in the first round of searching process. By integrating database searches, a chemical retrosynthetic pathway tree was recursively constructed.

When the agent encounters an intermediate node that cannot be expanded, it queries about five additional articles on its synthesis methods to extract supplementary chemical reactions, thereby helping to extend the reaction pathway to available compounds as initial reactants. In the end, the agent supplemented with 158 additional papers on intermediate synthesis reactions, processed a total of 197 papers, and obtained an expanded chemical retrosynthetic pathway tree for polyimide (Fig.~\ref{fig:fig4}). Ultimately, the number of nodes in the Rretrosynthetic Pathway Tree increased from the original 322 to 3099, and the number of synthesis pathways identified through the MBRPS algorithm increased from 55 to 292.

\subsection{Evaluation and Recommendation for Chemical Synthesis Pathways}

Most studies for retrosynthesis planning focus solely on reactants and products, neglecting reaction conditions.\cite{Segler2018,Zhao2024,Chen2020,Lin2020,Karpov2019} However, factors such as reaction mildness, reactant availability and cost, yield and scalability, and safety profile are crucial considerations in retrosynthesis planning. 
Due to the large number of obtained reaction pathways, the agent initially screens reactions within the retrosynthetic pathway tree based on these conditions (see Section III of SI for details). 

Finally, the agent employs Chain-of-Thought (CoT)\cite{Wei2022} reasoning to conduct a comprehensive evaluation of each reaction pathway that has passed the initial screening and been validated. This evaluation considers each pathway’s advantages and disadvantages based on the specific criterion designated by humans. In practical applications, the recommendation criteria can be adjusted based on specific needs. we provide the following two criteria for demonstration purposes:

1. Method for producing commercially available Katpon polyimide.

2. Presence of specific compounds in the initial reactants.

\begin{figure}[h]
\centering
\includegraphics[width=0.8\linewidth]{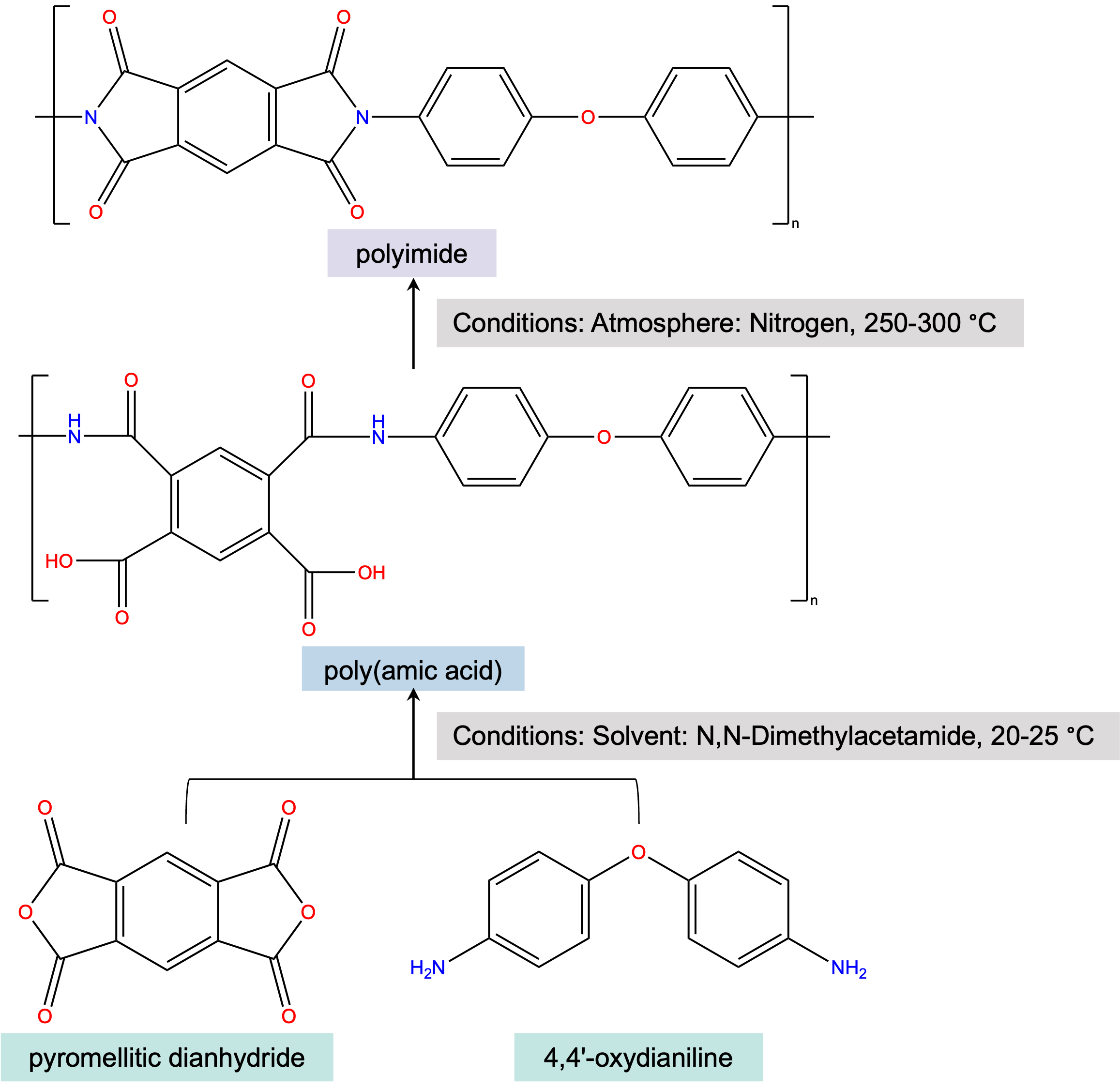}
\caption{The optimal reaction pathway recommended by agent based on criterion 1 (Also see Fig.~\ref{fig:fig4})}
\label{fig:fig5}
 \end{figure}

\begin{figure*}[h]
\centering
\includegraphics[width=0.85\linewidth]{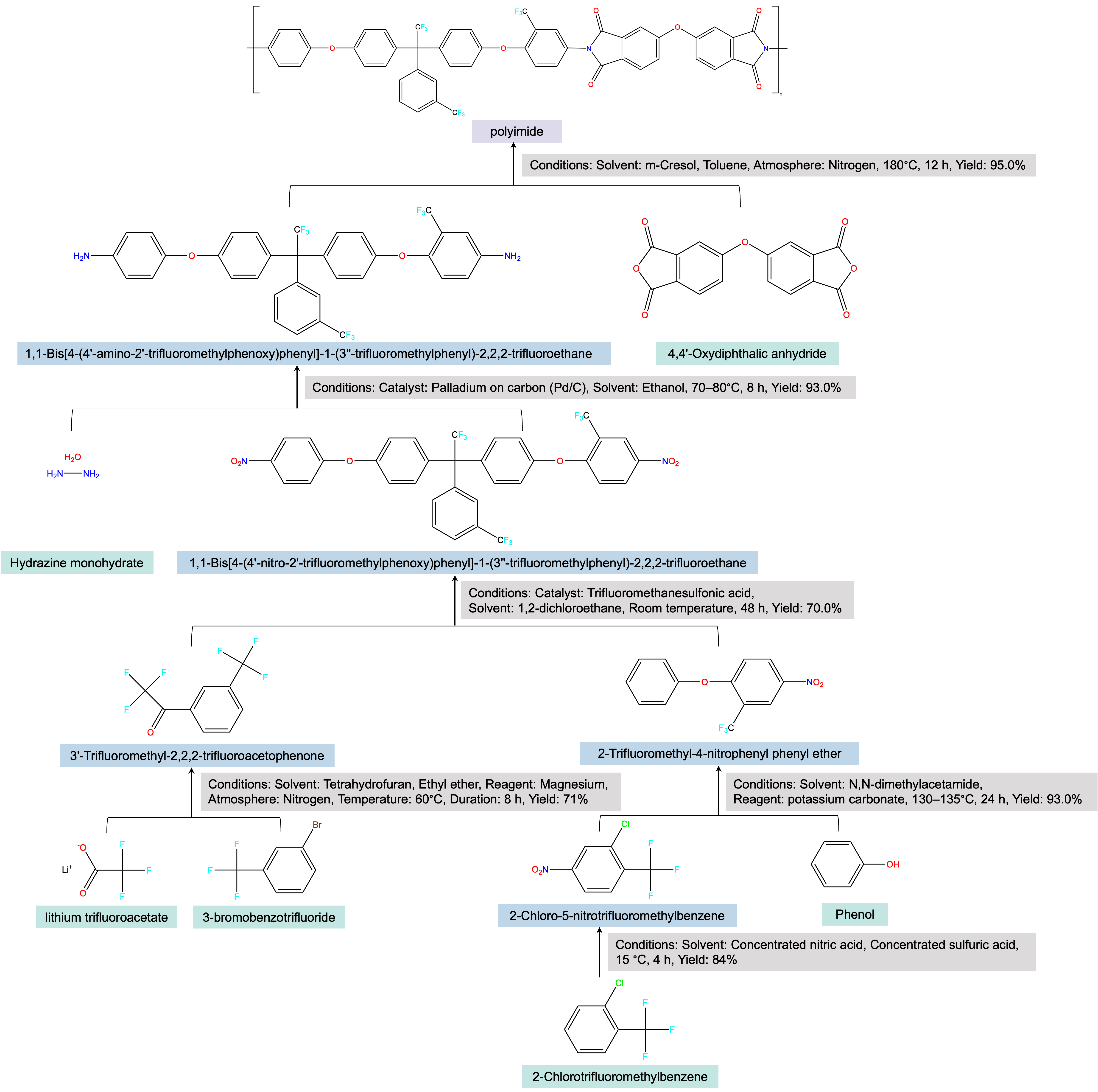}
\caption{The optimal reaction pathway recommended by agent based on criterion 2 (Also see Fig.~\ref{fig:fig4}).}
\label{fig:fig6}
\end{figure*}

Based on this detailed evaluation, the agent then recommends the optimal reaction pathway, along with a rationale explaining how it best the outlined criteria. The final recommended reaction pathways are presented in Fig.~\ref{fig:fig5} (based on Criterion 1) and Fig.~\ref{fig:fig6} (based on Criterion 2) (see Section 4 of SI for details). Notably, the reaction pathway obtained based on Criterion 2 is one of the newly proposed pathways. It was identified by the agent through an extended search of the literature related to intermediate synthesis. This approach enables the discovery of additional alternative pathways to better meet the demands of various practical application scenarios.

\section{Discussion}

\subsection{Challenges and Proposed Solutions in Macromolecule Retrosynthesis Planning}

Information extraction for polymer materials presents greater challenges compared to small-molecule chemicals. Small molecules benefit from standardized representations such as SMILES (Simplified Molecular Input Line Entry System)\cite{Weininger1988,Weininger1989} and IUPAC (International Union of Pure and Applied Chemistry) nomenclature, which provide unique and structured identifiers for molecular structures. In contrast, polymers lack a single, universally recognized naming standard. Their nomenclature often varies based on their naming systems, monomer composition, topology, material properties, application scenarios, and other factors. For instance, Poly(vinyl acetate) (PVA) can be named "Poly(1-acetoxyethylene)" or "Poly(ethenyl acetate)" based on different nomenclature systems.\cite{Hodge2020,Hu2021} Similarly, polyimides (PI) can be named "poly(amide-imide)" or "poly(1,3-dioxoisoindoline-2-yl) ethylene" based on their structural characteristics.\cite{Xu2021}

Traditional rule-based methods struggle with these complexities, while LLMs excel at distinguishing various chemical substance names and accurately extracting polymer-related information, without relying on a predefined format. However, due to input length limitations, LLMs ensure consistency only within single articles. To address cross-article inconsistency, the agent reviews the knowledge graph, unifies duplicate nodes, and corrects the graph to achieve entity alignment.

With the help of LLMs and specifically designed techniques, the names of the extracted chemical compounds from different articles were aligned to a great extent, ultimately leading to the completion of the PI reaction pathway tree (Fig.~\ref{fig:fig4}). However, a very small number of duplicate compounds inevitably remained due to the limitations of LLMs, which have been corrected. This minor error highlights the challenges posed by the naming conventions of polymer systems. Furthermore, the large-scale application of our method relies on future improvements in the ability of LLMs to accurately identify chemically identical compounds with different names without omissions.

\subsection{Advantages of Using Knowledge Graphs in Macromolecule Retrosynthesis Planning}

When processing large volumes of academic literature, traditional Retrieval-Augmented Generation (RAG) techniques\cite{Lewis2020} are helpful in reducing hallucinations by linking answers to retrieved documents. However, they face significant limitations, including poor document retrieval quality, suboptimal ranking of relevant documents, and unstructured data management.\cite{Lewis2020,Barnett2024,Gao2023} These shortcomings often lead to incomplete or misleading responses, particularly in retrosynthesis planning, where precision is paramount.

To address these issues, we adopt a structured knowledge graph to store information on chemical reactions from various sources, rather than relying on the vector-based retrieval mechanism in RAG, which typically retrieves information from unstructured text embeddings. By leveraging the knowledge graph, agents can accurately and efficiently retrieve data to construct retrosynthetic pathway trees.  This method is highly scalable, allowing agents to explore relevant synthesis literature and extend intermediates to leaf nodes for reactions that cannot be expanded. It also supports dynamic updates by integrating the latest academic papers, effectively mitigating the knowledge update lag in LLMs. Each chemical reaction is paired with a literature reference, addressing issues of hallucination and unverifiability in LLMs. This enhances the accuracy, reliability, and authority of reaction pathway recommendations.

\subsection{Key Advantages of Our Method for Macromolecule Retrosynthesis Planning}
The current methods for single-step chemical retrosynthesis analysis (predicting reactants based on a given product) primarily include utilizing deep learning models for prediction\cite{Zhao2024,Chen2020,Lin2020,Karpov2019} and employing density functional theory (DFT) for precise calculations\cite{Li2024}. These methods are generally limited to the study of small chemical molecules, mainly due to the lack of databases on polymer chemical reactions, the large number of atoms in macromolecular systems (typically on the order of \(10^{2} - 10^{6}\))\cite{Kremer1990}, and the fact that chemical reactions often involve long-range interactions in macromolecules and solvent effects, making accurate calculations challenging. To address these limitations, we propose a novel and practical approach that employs an LLM agent using authoritative academic papers as the knowledge source to perform multi-step chemical retrosynthesis analysis for polymer materials.

Our method stands out for its high interpretability and reliability, as it is grounded in experimental validation from authoritative academic papers. In comparison, template-free deep learning models for single-step retrosynthesis struggle with relatively low prediction accuracy (around 40-60\%)\cite{Lin2020,Karpov2019}, making it challenging to generate complete and valid pathways. Although template-based deep learning methods achieve higher accuracy (approximately 70-100\%)\cite{Zhao2024,Chen2020}, they rely heavily on predefined annotated reaction templates, limiting their flexibility. In contrast, our approach not only provides highly accurate and valid reaction pathways for polymer materials, such as polyimides, with accuracy estimated to be in the high 90s, validated by databases and traceable source literature, but also offers multiple viable pathways tailored to different application needs, thereby enhancing practical value in retrosynthesis planning.
Additionally, the vast majority of these methods are based on a "one-to-one" decomposition strategy (where a product is decomposed into at most one reaction intermediate), resulting in unbranched reaction pathways that facilitate search using Monte Carlo Tree Search (MCTS). In practical scenarios, however, "one-to-more" decomposition strategies (where a product decomposes into one or more reaction intermediates) are more common, leading to multi-branched reaction pathways. To better align with practical application scenarios, we utilize the Memoized Depth-first Search (MDFS) algorithm to construct a retrosynthetic pathway tree based on a knowledge graph and employ the Multi-branched Reaction Pathway Search algorithm (MBRPS) algorithm to identify all possible reaction pathways, specifically designed for multi-branched retrosynthetic pathways. This approach enables the identification of all viable reaction pathways, providing all necessary reactions (including reaction conditions) starting from available chemical compounds as initial reactants to synthesize the target product.

\section{Conclusion}
This study represents the first attempt to develop a fully automated retrosynthesis planning agent  specifically designed for macromolecules by integrating large language models with knowledge graphs. Demonstrated through a case study on polyimide, the approach automates literature retrieval, reaction data extraction, database querying, construction of retrosynthetic pathway trees, further expansion through the retrieval of additional literature on intermediates, finally searching, evaluation and recommendation of the optimal route based on conditions, reactants, safety, and other factors. Our approach is versatile and not limited to small molecules but extends to complex macromolecules. In contrast to previous methods that have been limited to "one-to-one" decomposition strategy, our method is suitable for "one-to-many" decomposition strategy, a scenario more commonly encountered in practical chemical synthesis analysis. By applying this approach to the widely-used polyimide, the agent successfully constructs the retrosynthesis pathway tree, and recommend both established and novel pathways without human intervention. This example demonstrates that with more powerful LLMs, an automated retrosynthesis planning agent could significantly accelerate the discovery of reverse chemical reaction pathways, thereby greatly enhancing research efficiency.


\section*{Code availability}
The source code of RetroSynthesisAgent is available at \href{https://github.com/QinyuMa316/RetroSynthesisAgent}{https://github.com/QinyuMa316/RetroSynthesisAgent}, where we provide a demo video of its usage.

\section*{Supporting Information}
The Supporting Information is available free of charge at
A demo video demonstrating the operation process of the automated retrosysthesis planning (MP4).

\section*{Conflicts of interest}
There are no conflicts to declare.

\section*{Acknowledgements}
We gratefully acknowledge Prof. Junpo He (Fudan University), a polymer chemist, for his valuable insights and discussions. This work was supported by grants from the National Natural Science Foundation of China (Nos. 22373022, 52394272), the National Key Research and Development Program of China (No. 2023YFA0915300), and the Shanghai Science and Technology Innovation Action Plan (No. 24JD1400700).



\balance



\bibliographystyle{rsc} 
\bibliography{Retrosynthesis_LLM} 

\end{document}